\documentclass[lettersize,journal,twoside]{IEEEtran}
\usepackage{amsmath,amsfonts,amssymb,bm}
\usepackage[linesnumbered,ruled,vlined]{algorithm2e}
\usepackage{algpseudocode}
\usepackage{array}
\usepackage{textcomp}
\usepackage{stfloats}
\usepackage{url}
\usepackage{graphicx} \graphicspath{{images/}}
\usepackage{balance}
\usepackage[caption=false,font=footnotesize]{subfig}
\usepackage{placeins}
\usepackage{pifont}
\usepackage{kotex}
\usepackage{xcolor}
\usepackage{pdfpages}

\newcommand{\bmf}[1]{\bm{\mathrm{#1}}}

\begin{document}
\includepdf[pages=1]{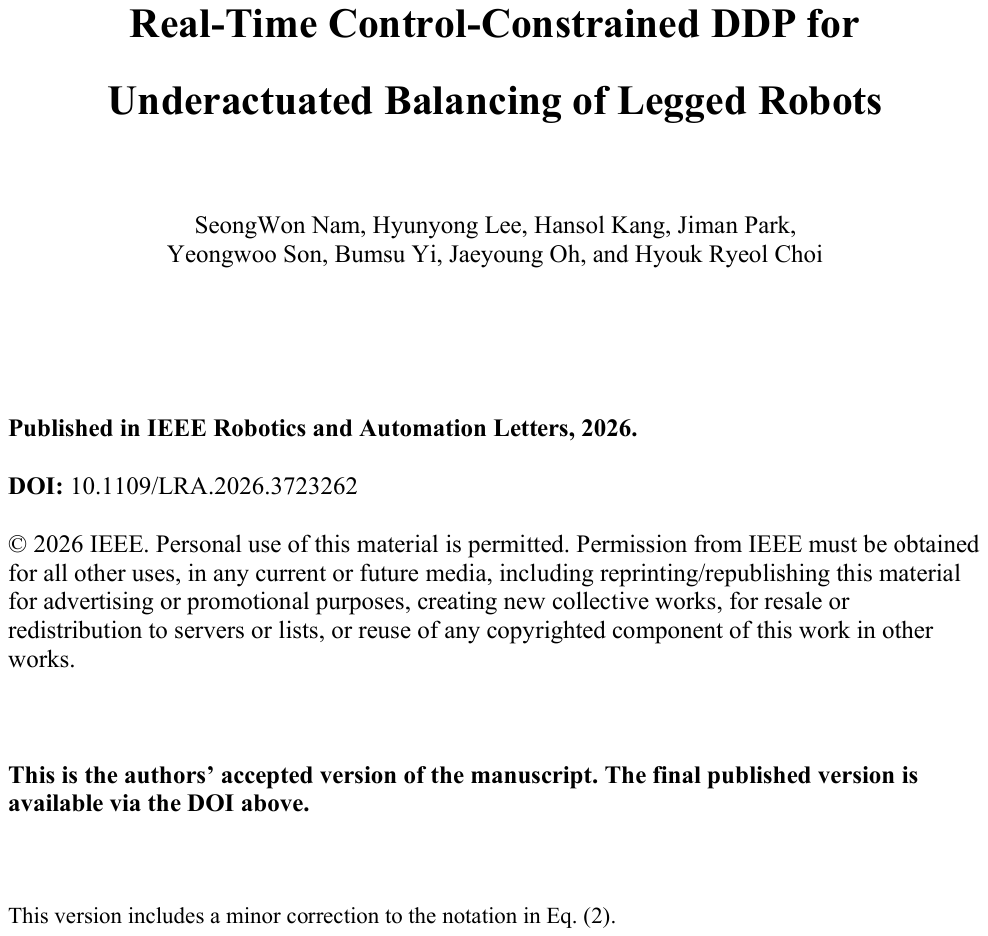}

\title{ Real-Time Control-Constrained DDP for Underactuated Balancing of Legged Robots }

\author{
    SeongWon~Nam$^{1}$,
    Hyunyong~Lee$^{2}$,
    Hansol~Kang$^{1}$,
    Jiman~Park$^{1}$,
    Yeongwoo~Son$^{1}$, \\
    Bumsu~Yi$^{1}$, 
    Jaeyoung~Oh$^{1}$,
    and Hyouk~Ryeol~Choi$^{1,2}$ \IEEEmembership{Fellow, IEEE}

    \thanks{Manuscript received: March 31, 2026; Accepted: July 20, 2026.}
    \thanks{This paper was recommended for publication by Editor Lucia Pallottino upon evaluation of the Associate Editor and Reviewers comments.}
    \thanks{This work was supported by the Materials and Parts Technology Development Program 
            (RS-2024-00508191, Development and Demonstration of Unmanned Autonomous Operation Technology Based on Field-Use Visualization Sensors and 6-Axis Rotational Angle Sensors) 
            funded by the Ministry of Trade Industry \& Energy (MOTIE, Korea) (\textit{Corresponding author: Hyouk Ryeol Choi}).}
    \thanks{$^{1}$Robotics Innovatory, School of Mechanical Engineering, Sungkyunkwan University (SKKU), South Korea
            (e-mail: sholybest@g.skku.edu, choihyoukryeol@gmail.com).}
    \thanks{$^{2}$AIDIN ROBOTICS Inc., Anyang, South Korea.}
    \thanks{Digital Object Identifier (DOI): 10.1109/LRA.2026.3723262}
}

\markboth{IEEE Robotics and Automation Letters, preprint version, 2026}
{S. Nam \MakeLowercase{et al.}: Real-Time Control-Constrained DDP for Underactuated Balancing of Legged Robots}

\maketitle

\begin{abstract}
    This paper presents a real-time control-constrained Differential Dynamic Programming (DDP) framework for underactuated legged robots.
    To address the limitation of classical DDP in handling control constraints, we propose an Accelerated Projected Gradient (APG)-based control-constrained DDP (ABC-DDP), which efficiently computes constrained solutions and identifies active sets without repeated Karush–Kuhn–Tucker (KKT) inversions.
    A virtual constraint is introduced to integrate control constraints within a feasibility-driven multiple-shooting framework, enabling stable optimization even from dynamically infeasible initializations.
    The proposed method supports real-time model predictive control (MPC) with short horizons under strong underactuation.
    Simulation results demonstrate static two-leg standing under external disturbances, along with diverse dynamic motions including slow catwalk, upright walking, and high-speed running within a unified MPC framework.
    To the best of our knowledge, this is the first demonstration of static two-leg standing of a quadruped robot achieved using real-time finite-horizon MPC.
\end{abstract}


\section{Introduction} \label{sec: Introduction}
    \IEEEPARstart{M}{odel} Predictive Control has recently enabled remarkable performance in robotic locomotion. 
    By directly discretizing the optimal control problem and exploiting the recursive structure of the system dynamics, the resulting problem can be solved efficiently. 
    As shown in \cite{C-MPC,RF-MPC,HONG-NMPC,AnmalWBC2023TRO}, such approaches typically reduce to solving one or multiple Quadratic Programming (QP) problems that approximate the original problem with dynamics and control constraints.
    A key challenge in this framework is the proper handling of nonlinear dynamics constraints. 
    Differential Dynamic Programming (DDP) \cite{THE-FIRST-DDP} provides a structured way to incorporate dynamics constraints and achieves superlinear convergence in the vicinity of a reference trajectory. 
    However, classical DDP does not explicitly handle control constraints. 
    Notably, even simple control constraints, such as box constraints or friction cone constraints, are often sufficient to enable a wide range of locomotion tasks when properly handled.

    To handle simple box-type control limits within DDP, Box-DDP \cite{Box-DDP} has been proposed.
    By exploiting the box structure of the constraints, it constructs a \textit{free-space} Hessian by removing the rows and columns of the control hessian corresponding to the active box boundaries.
    Extending this approach directly to more general linear constraints introduces additional challenges.
    Rather than simply removing rows and columns, one would need to eliminate directions spanned by the normal vectors of the active constraint boundaries when forming the constrained hessian.
    This would likely require matrix factorizations at each sub-iteration to appropriately transform the coordinate system, leading to a significant loss in computational efficiency.
    Constrained DDP (CDDP) \cite{XieConstrainedDDP} instead formulates a Karush–Kuhn–Tucker (KKT) system that incorporates the control hessian and linearized active constraints within the DDP subproblem.
    The inversion of the control hessian already constitutes a substantial portion of the DDP computational cost.
    If an active-set QP method were employed to accurately identify the active set, it would require solving a KKT system at every sub-iteration, potentially increasing the computational burden by several factors.
    To mitigate this, CDDP adopts a practical strategy that first starts by evaluating whether the current trajectory is active in order to identify the active set.

    In this context, accelerating a projection-based first-order method can provide an effective alternative, as gradient-based iterations are computationally inexpensive while still enabling reliable identification of the active set.
    The Accelerated Projected Gradient (APG) method \cite{APG-NIPS2015}, originally developed in the context of image processing, has been shown to efficiently solve large-scale, dense constrained problems, and even certain nonconvex problems, without requiring hessian inversion.
    Although computing the projection itself can be viewed as an optimization problem, commonly used control constraints in legged robotics, such as box constraints and friction cone constraints, admit closed-form solutions, enabling fast and numerically reliable computation.
    From this perspective, APG is well suited for handling simple control constraints.
    However, it may be less effective for constraints whose projections are difficult to compute, such as nonlinear dynamics constraints.
    Nevertheless, when combined with DDP, which inherently satisfies dynamics constraints but does not explicitly handle control constraints, the two approaches can provide complementary advantages.

    Meanwhile, DDP seeks an optimal solution efficiently by starting from a nominal trajectory and applying second-order Taylor approximations around it.
    However, in scenarios where underactuation persists for an extended period, constructing a dynamically feasible trajectory from scratch can be challenging.
    If such a trajectory were readily available, constructing it would itself constitute a nontrivial trajectory optimization problem.
    Furthermore, as DDP is fundamentally based on a direct single-shooting formulation, integration errors can accumulate over time, which may lead to divergence when considering long horizons or highly dynamic trajectories with large fluctuations.
    Feasibility-driven DDP (FDDP) \cite{FDDP} addresses this issue by enabling multiple-shooting behavior through independent state-control sequences.
    Starting from a dynamically infeasible trajectory, it iteratively steers the solution toward a feasible one.
    Box-FDDP \cite{Box-FDDP}, which combines FDDP and Box-DDP, has also been proposed and has demonstrated brilliant performance in applications such as \cite{Box-FDDP-Locomotion, KimIJRRcontactImplicit}.
    However, since control limits and dynamic feasibility could not be handled simultaneously within a unified framework, these methods were applied in an alternating manner. 

    The main contributions of this work are as follows:
    \begin{itemize}
        \item Existing constrained DDP techniques have largely remained fragmented, each offering distinct advantages and limitations. 
              By incorporating the APG method, which has received limited attention in the control community, 
              the proposed framework extends beyond the box-constrained setting of Box-DDP and the indirect active-set treatment of CDDP, 
              enabling efficient handling of linear control constraints with exact active-set identification.
        \item We develop a unified control-constrained multiple-shooting DDP framework. 
              While Box-DDP and FDDP alternatively address box-constrained mode and unconstrained multiple-shooting mode in Box-FDDP, no existing method simultaneously considers control constraints and a multiple-shooting scheme. 
              This is achieved by introducing a novel \textit{virtual constraint} formulation.
        \item The proposed framework enables static two-leg standing of a quadruped, while also demonstrating a variety of dynamic locomotion and balancing tasks within the same framework.
    \end{itemize}

\begin{figure}[!t]
        \centering
        \includegraphics[width=1.0\columnwidth]{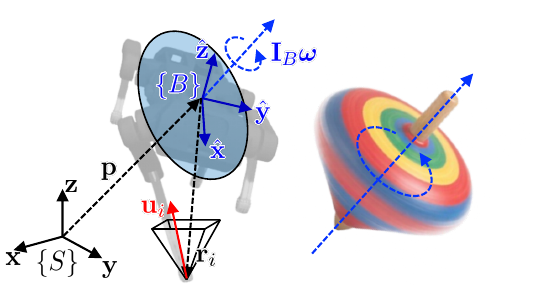}
        \caption{
            SRBD representation of a quadruped robot ($\{S\}$: inertial frame, $\{B\}$: body frame). 
            The body motion induced by ground reaction forces is illustrated together with an analogy to a spinning top, highlighting precession and nutation effects.
        }
        \label{fig: SRBD}
    \end{figure}

    The \textit{contact-implicit} MPC formulation that explicitly incorporates contact dynamics was presented in \cite{KimIJRRcontactImplicit} and solved using Box-FDDP, demonstrating two-leg balancing behavior. 
    However, the resulting motions did not exhibit static standing with fixed foot contacts, as the optimizer simultaneously planned separating, clamping, and sliding contact phases. 
    Note that two-leg standing in a point-foot quadruped corresponds to balancing on a support line (not a polygon), analogous to a human balancing on tiptoes without taking steps. 
    As illustrated in Fig.~\ref{fig: SRBD}, maintaining such a posture generally requires complex motion planning that exploits rotational dynamics, including precession and nutation effects.
    The practical feasibility of two-leg standing was also demonstrated in \cite{VariationalQP-2LEGS}, where a single-step QP was derived by approximating the constrained cost-to-go (CCTG) term in the continuous Hamilton–Jacobi–Bellman (HJB) equation with its unconstrained counterpart (UCTG). 
    Nevertheless, MPC-based approaches were considered challenging to implement in real time due to the need for long-horizon prediction.
    Despite this common belief, this work shows that underactuated conditions can be sustained even with a short finite-horizon prediction.
    To the authors’ knowledge, this is the first demonstration of static two-leg standing of a quadruped robot realized within a real-time finite-horizon MPC framework.
    In addition, the proposed framework generates a diverse set of dynamic motions under strong underactuation, including slow catwalk, upright bipedal walking, and high-speed running at 6.0~m/s, all within a single MPC framework in real-time simulation. 

\section{MPC Formulation} \label{sec: MPC}  

    MPC problem is formulated as:
    \begin{equation} \label{eq: MPC-msform}
        \begin{aligned}
            \min_{\substack{\bmf{x}_1,\ldots,\bmf{x}_N \\ \bmf{u}_0,\ldots,\bmf{u}_{N-1}}}
                \sum_{k=0}^{N-1} & \ell(\bmf{x}_k,\bmf{u}_k) + \ell_N(\bmf{x}_N) \\
            \mathrm{s.t.} \quad &\bmf{x}_{k+1} = \bmf{f}_k(\bmf{x}_k,\bmf{u}_k), \hspace{0.5em} \bmf{x}_0 = \tilde{\bmf{x}}_0 \\
                                &\bmf{u}_k \in C_k.
        \end{aligned}
    \end{equation} 

    \noindent
    with states $\bmf{x}_k$, control input $\bmf{u}_k$, and discretized system dyna-mics $\bmf{f}_k$.
    $\tilde{\bmf{x}}_0$ means a measured initial state.
    Each stage cost $\ell$ is defined by a sum of weighted quadratic costs for a vectorized state error metric $\bmf{\xi}_k$ 
    and diagonal weight matrices $\bmf{W}$'s:
    \begin{equation}
        \begin{aligned}
            \ell(\bmf{x}_k,\bmf{u}_k) &:= \frac{1}{2}\left\|\bm{\bmf{\xi}}_k({\bmf{x}_k})\right\|_{\bmf{W}_{\bmf{x}_k}}^2 
            + \ell_{\bmf{u}_k}(\bmf{u}_k), \\
            \ell_{\bmf{u}_k}(\bmf{u}_k) &:= \frac{1}{2}\left\|\bmf{u}_k\right\|_{\bmf{W}_{\bmf{u}_k}}^2 
            + \frac{1}{2}\left\|\bmf{u}_k-\overline{\bmf{u}}_{k+1}\right\|_{\bmf{W}_{\Delta\bmf{u}_k}}^2,
        \end{aligned}
    \end{equation}
    where $\overline{\bmf{u}}_{k+1}$ is taken from the current nominal trajectory.
    In this work, the system dynamics is simplified to a Single Rigid Body Dynamics (SRBD) model, as shown in Fig.~\ref{fig: SRBD},
    for real-time computational efficiency in underactuated conditions.
    The control input $\bmf{u}_k$ consists of the ground reaction forces (GRFs) generated by each leg.
    The system dynamics, error parametrization and associated derivatives follow the variational approach presented in \cite{HONG-NMPC}.

    The control constraint set $C_k$ represents the linear friction pyramid and box constraints.
    Specifically, for the GRF from the $i^{\mathrm{th}}$ leg and the friction coefficient $\mu$,
    \begin{equation} \label{eq: pyramid}
        \lvert u^{i}_{kx} \rvert \le \mu u^{i}_{kz}, \hspace{0.5em}
        \lvert u^{i}_{ky} \rvert \le \mu u^{i}_{kz}, \hspace{0.5em}
        u_{z,min} \le u^{i}_{kz} \le u_{z,max}.
    \end{equation}
    For the legs in swing phase, $u_{z,min} = u_{z,max} = 0$.

\section{Accelerated Projected Gradient Method} \label{sec: APG}
    To obtain the constrained minimum and identify the active set in the DDP subproblem, we employ the Nonmonotone Accelerated Projected Gradient (APG) method proposed in \cite{APG-NIPS2015}.
    Consider the following constrained optimization problem for a real scalar function $F$ and $\bmf{x} \in \mathbb{R}^n$:
    \begin{equation} \label{eq: constrained-optimization}
        \min_{\bmf{x}\in\mathbb{R}^n} \quad F(\bmf{x}) \qquad \mathrm{s.t.} \quad \bmf{x} \in O.
    \end{equation}
    The projected gradient method performs the following update at each iteration $k$:
    \begin{equation} \label{eq: proj-grad}
        \begin{aligned}
            \bmf{x}_{k+1} &= P_O(\bmf{x}_k - \alpha_{k}\nabla{F(\bmf{x}_k)}), \\
            P_O(\bmf{x}) &:= \operatorname{argmin}_{\bmf{o} \in O} \| \bmf{o}-\bmf{x} \|_2
        \end{aligned}
    \end{equation}
    where $\alpha_k$ is the step size and $P_O$ is the orthogonal projection onto the constraint set $O$ as defined in \cite{Fisrt-Order-Book}. 
    To determine the step size, we employ backtracking line search with the Barzilai–Borwein (BB) method \cite{Barzilai-Borwein-Rule}.
    The BB rule provides a good initial guess for the step size by capturing the curvature trend between the two most recent iterates.
    For $0 < \rho < 1$, the BB step size $\alpha$ is obtained by approximating $F(\alpha)$ along the search direction using a second-order model:
    \begin{equation} \label{eq: BB-shortstep}
        \begin{aligned}
            \alpha_0 &= \operatorname{argmin}_{\alpha} \| \Delta\bmf{x} - \alpha\Delta\bmf{g} \|^2 = 
             \frac{\Delta\bmf{x}^\top\Delta\bmf{g}}{\Delta\bmf{g}^\top\Delta\bmf{g}}, \\
            \alpha_k &= \rho \cdot \alpha_{k-1} \,\, (k\in\{1, 2, 3, \cdots\})
        \end{aligned}
    \end{equation}
    where $\Delta\bmf{x}$ and $\Delta\bmf{g}$ denote the changes in the variable and gradient, respectively.
    During line search, enforcing strict monotonic decrease can result in overly conservative step sizes, particularly for ill-conditioned problems.
    To alleviate this, APG employs the following relaxed objective value $c_k$ in its line search acceptance criterion:

    \begin{equation} \label{eq: ck}
        c_k = \frac{\sum_{j=1}^{k}\eta^{k-j}F(\bmf{x}_j)}{\sum_{j=1}^{k}\eta^{k-j}}
    \end{equation}
    where $\eta \in [0,1)$ serves as the controlled degree of nonmonotonicity.
    The value of $c_k$ can be updated efficiently using the following recursion:
    \begin{equation} \label{eq: ck-iterative}
    \begin{aligned}
        q_{k+1} &= \eta q_k + 1, \\
        c_{k+1} &= \frac{\eta q_k c_k + F(\bmf{x}_{k+1})}{q_{k+1}}.
    \end{aligned}
    \end{equation}

    \begin{algorithm}[!t]
        \caption{APG with backtracking line search \cite{APG-NIPS2015}}\label{alg: APG}
        \textbf{Initialize} $t_0=0$, $t_1=1$, $q_1=1$, $\eta\in[0,1)$, $\delta>0$, $\rho<1$, 
        $c_1=F(\bmf{x}_0)$, $\bmf{y}_1=\bmf{x}_0$, $\bmf{y}_0=\bmf{0}$, $\nabla F(\bmf{y}_0)=\bmf{0}$ \\[0.5ex]
        \For{$k = 1,2,3, \cdots$}{
            Compute $\nabla F(\bmf{y}_k)$ and initialize $\alpha$ by \eqref{eq: BB-shortstep} \label{algline: PGD repeat start} \\

            \Repeat{$ \left( c_k - F(\bmf{z}_{k+1}) \geq \delta \|\bmf{y}_k - \bmf{z}_{k+1}\|^2 \right) $}{
                $\bmf{z}_{k+1} = P_O( \bmf{y}_k-\alpha\nabla F(\bmf{y}_k) )$ \\[0.5ex]
                $\alpha=\rho\cdot\alpha$ \\[0.5ex]
            } \label{algline: PGD repeat end}
            \vspace{1.0ex}
            \If{ $ \left( c_k - F(\bmf{z}_{k+1}) \geq \delta \|\bmf{y}_k - \bmf{z}_{k+1}\|^2 \right) $ }{
                $\bmf{x}_{k+1} = \bmf{z}_{k+1}$
            }
            \Else{
                Repeat \ref{algline: PGD repeat start} -- \ref{algline: PGD repeat end},
                with $\bmf{y}_{k}$ and $\bmf{z}_{k+1}$ replaced by $\bmf{x}_{k}$ and $\bmf{v}_{k+1}$, respectively.

                \vspace{0.5ex}
                $
                    \bmf{x}_{k+1} = \begin{cases}
                        \bmf{z}_{k+1}, & \text{if } F(\bmf{z}_{k+1}) \leq F(\bmf{v}_{k+1}), \\
                        \bmf{v}_{k+1}, & \text{otherwise.}
                    \end{cases}
                $
            }
            \vspace{0.5ex}
            Update $q_{k+1}$ and then $c_{k+1}$ by \eqref{eq: ck-iterative}. \\
            \vspace{0.5ex}
            \If{$|c_{k}-c_{k+1}|^2 < \varepsilon$}{
                \textbf{break}
            }
            Update $t_{k+1}$ and then $\bmf{y}_{k+1}$ by \eqref{eq: Nesterov-update}.
        }
    \end{algorithm}

    \noindent
    A larger value of $\eta$ gives more weight to the earlier (potentially larger) objective values in the iteration history, 
    making it easier for $F$ in later iterates to fall below $c_k$.
    This improves convergence in ill-conditioned settings and also ensures convergence for even nonconvex problems.
    As a result, the DDP subproblem can be solved within the feasible set even when the problem is not strictly positive definite.

    Furthermore, by employing a search direction based on Nesterov momentum \cite{Nesterov}, a convergence rate of $O(1/k^2)$ can be achieved in locally convex regions.
    The Nesterov-type momentum step $\bmf{y}$ is updated according to the following rule:
    \begin{equation} \label{eq: Nesterov-update}
        \begin{aligned}
            t_{k+1}       &= \dfrac{1+\sqrt{1+4t_{k}^2}}{2}, \\
            \bmf{y}_{k+1} &= \bmf{x}_{k} + \dfrac{t_k}{t_{k+1}}(\bmf{z}_{k}-\bmf{x}_{k}) + \dfrac{t_k-1}{t_{k+1}}(\bmf{x}_{k}-\bmf{x}_{k-1}). 
        \end{aligned}
    \end{equation}
    The overall procedure of APG is summarized in Algorithm~\ref{alg: APG}.

\section{APG-Based Control-Constrained DDP} \label{sec: ABCDDP}
    According to Bellman's Principle of Optimality \cite{Bellman-DP}, once the state at time $k$ is determined, the optimal future trajectory is independent of past decisions.
    Let $V_k(\bmf{x}_k)$ denote the value function (i.e., the optimal cost-to-go) at time $k$:
    \begin{equation}
        V_k(\bmf{x}_k) = \min_{\substack{ \bmf{u}_k, \cdots, \bmf{u}_{N-1} }} \sum_{j=k}^{N-1} \ell(\bmf{x}_j,\bmf{u}_j) + \ell_N(\bmf{x}_N)
    \end{equation}
    This can be reformulated as follows based on the optimality principle:
    \begin{equation} \label{eq: BellmanEqn}
        V_k(\bmf{x}_k) = \min_{\bmf{u}_k} \ell(\bmf{x}_k,\bmf{u}_k) + V_{k+1}(\bmf{f}_k(\bmf{x}_k,\bmf{u}_k)).
    \end{equation}
    If the value function at time $k+1$ is known, \eqref{eq: BellmanEqn} shows that the value function at time $k$ can be expressed solely in terms of the variables at step $k$.
    By specifying the boundary condition $V_N(\bmf{x}_N) = \ell_N(\bmf{x}_N)$, the value function can be computed recursively in a backward manner starting from step $N-1$, which is referred to as the backward pass.
    Furthermore, the argument of \eqref{eq: BellmanEqn} is referred to as the action-value function and is denoted by $Q$ as follows:
    \begin{equation}
    \begin{aligned}
        Q_k(\bmf{x},\bmf{u}) &= \ell_k(\mathbf{x}, \mathbf{u}) + V_{k+1}(\bmf{f}_k(\mathbf{x}, \mathbf{u})) \\
        &= Q_k(\bmf{x}_k+\delta\bmf{x}_k,\bmf{u}_k+\delta\bmf{u}_k)\\
        &\approx Q_k(\bmf{x}_k,\bmf{u}_k) + \delta{Q}_k(\delta\bmf{x}_k,\delta\bmf{u}_k)
    \end{aligned}
    \end{equation}
    The goal of DDP backward pass is to minimize \eqref{eq: Q-Taylor}, which is obtained by applying a second-order Taylor approximation of the Q-function around the current trajectory:
    \begin{equation} \label{eq: Q-Taylor}
    \begin{aligned}
        \delta{Q}_k(\delta\bmf{x}_k,\delta\bmf{u}_k) = 
        \frac{1}{2} \hspace{-0.35em}
        \begin{bmatrix}
        1 \\
        \delta\mathbf{x}_k \\
        \delta\mathbf{u}_k
        \end{bmatrix}^{\hspace{-0.35em}\top} \hspace{-0.6em}
        \setlength{\arraycolsep}{2.5pt}
        \begin{bmatrix}
        0 & Q_{\mathbf{x}_k}^{\top} & Q_{\mathbf{u}_k}^{\top} \\
        Q_{\mathbf{x}_k} & Q_{\mathbf{x}\mathbf{x}_k} & Q_{\mathbf{x}\mathbf{u}_k} \\
        Q_{\mathbf{u}_k} & Q_{\mathbf{u}\mathbf{x}_k} & Q_{\mathbf{u}\mathbf{u}_k}
        \end{bmatrix} \hspace{-0.45em}
        \begin{bmatrix}
        1 \\
        \delta\mathbf{x}_k \\
        \delta\mathbf{u}_k
        \end{bmatrix}
    \end{aligned}
    \end{equation}
    
    \noindent
    and the coefficients can be written as follows:
    \begin{subequations} \label{eq: Qdiffs}
    \begin{align}
        Q_{\mathbf{x}_k}
        &= \ell_{\mathbf{x}_k} + \mathbf{f}_{\mathbf{x}_k}^{\top} V_{\bmf{x}_{k+1}}, \\
        Q_{\mathbf{u}_k}
        &= \ell_{\mathbf{u}_k} + \mathbf{f}_{\mathbf{u}_k}^{\top} V_{\bmf{x}_{k+1}}, \\
        Q_{\mathbf{x}\mathbf{x}_k}
        &= \ell_{\mathbf{x}\mathbf{x}_k}
        + \mathbf{f}_{\mathbf{x}_k}^{\top} V_{\bmf{x}\bmf{x}_{k+1}} \mathbf{f}_{\mathbf{x}_k}
        + {V_{\bmf{x}_{k+1}} {\cdot\,} \mathbf{f}_{\mathbf{x}\mathbf{x}_k}}, \\
        Q_{\mathbf{u}\mathbf{u}_k}
        &= \ell_{\mathbf{u}\mathbf{u}_k}
        + \mathbf{f}_{\mathbf{u}_k}^{\top} V_{\bmf{x}\bmf{x}_{k+1}} \mathbf{f}_{\mathbf{u}_k}
        + {V_{\bmf{x}_{k+1}} {\cdot\,} \mathbf{f}_{\mathbf{u}\mathbf{u}_k}}, \\
        Q_{\mathbf{u}\mathbf{x}_k}
        &= \ell_{\mathbf{u}\mathbf{x}_k}
        + \mathbf{f}_{\mathbf{u}_k}^{\top} V_{\bmf{x}\bmf{x}_{k+1}} \mathbf{f}_{\mathbf{x}_k}
        + {V_{\bmf{x}_{k+1}} {\cdot\,} \mathbf{f}_{\mathbf{u}\mathbf{x}_k}} .
    \end{align}
    \end{subequations}
    In this work, an iterative Linear-Quadratic Regulator (iLQR) formulation is adopted, where second-order derivatives of the dynamics are neglected, and the hessian of the quadratic cost $\ell$ is Gauss--Newton approximated.
    
    \vspace{-0.6ex}
    \subsection{Constrained Backward Pass}
    As described earlier, the backward pass proceeds from the final time step and, at each step $k$, minimizes $\delta Q_k$:
    \begin{equation} \label{eq: constrainedDDP}
        \begin{aligned}
            \min_{\delta \mathbf{u}_k} \delta{Q_k} = 
            \frac{1}{2}\delta \mathbf{u}_k^{\top} &Q_{\mathbf{uu}_k} \delta \mathbf{u}_k
            +Q_{\mathbf{u}_k} \, \delta \mathbf{u}_k 
            + \delta\mathbf{x}_k^\top Q_{\mathbf{ux}_k}^\top \delta \mathbf{u}_k \\
            &\mathrm{s.t.} \ \bmf{A}_k\delta\bmf{u}_k - \bmf{b}_k \leq \bmf{0},
        \end{aligned}
    \end{equation}

    \noindent 
    where $\bmf{A}_k\delta\bmf{u}_k - \mathbf{b}_k \leq \bmf{0}$ characterizes the feasible region of $C_k$ defined in \eqref{eq: pyramid}.
    For the active constraints $\bmf{A}_k^*\delta\bmf{u}_k = \bmf{b}_k^*$, the following KKT system can be constructed \cite{XieConstrainedDDP}:
    \begin{equation} \label{eq: KKT-DDP}
        \begin{bmatrix}
            Q_{\mathbf{u}\mathbf{u}_k}  &  \bmf{A}_k^{*\top} \\[1ex]
            \bmf{A}_k^* &  \bmf{0} 
            \end{bmatrix}
            \begin{bmatrix}
            \delta \mathbf{u}_k \\[1ex]
            \bmf{\lambda} 
        \end{bmatrix}
        = -
        \begin{bmatrix}
            Q_{\mathbf{u}\mathbf{x}_k}\\[1ex]
            \bmf{0}   
        \end{bmatrix}
        \delta \mathbf{x}_k
        +
        \begin{bmatrix}
            -Q_{\bmf{u}_k}\\[1ex]
            \bmf{b}_k^*
        \end{bmatrix}
    \end{equation}
    where $\bmf{\lambda}$ denotes the Lagrange multiplier.
    It can be observed that this formulation is a constrained extension of the classical DDP backward pass.
    By solving this linear system, the optimal control input can be expressed as the sum of a feedback term and a feedforward term as follows:
    \begin{equation} \label{eq: backpass-solution}
        \delta\mathbf{u}_k^* = \bmf{K}_k\delta \bmf{x}_k + \bmf{k}_k
    \end{equation}
    where $\bmf{K}_k$ is called the feedback gain.
    Substituting this into the action-value function yields the value function at time $k$, allowing the backward pass to be carried out recursively.
    Taking derivatives of the resulting $V_k$ at $\bmf{x}_k$,
    \begin{equation} \label{eq: value-function}
    \begin{aligned}
        V_{\bmf{x}_k} &= 
        Q_{\bmf{x}_k}
        +
        \mathbf{K}_k^{\top}Q_{\bmf{uu}_k}\mathbf{k}_k
        +
        Q_{\bmf{ux}_k}^{\top}\mathbf{k}_k
        +
        \mathbf{K}_k^{\top}Q_{\bmf{u}_k}, \\
        V_{\bmf{xx}_k} &=
        Q_{\bmf{xx}_k}
        +
        \mathbf{K}_k^{\top}Q_{\bmf{uu}_k}\mathbf{K}_k
        +
        Q_{\bmf{ux}_k}^{\top}\mathbf{K}_k
        +
        \mathbf{K}_k^{\top}Q_{\bmf{ux}_k}.
    \end{aligned}
    \end{equation}
    A remaining question is how to identify the active set.
    It can be observed from \eqref{eq: KKT-DDP} that the same KKT matrix inversion is required for both the feedback and feedforward term.
    Moreover, the feedforward solution is equivalent to solving \eqref{eq: constrainedDDP} with $\delta\bmf{x}_k=\bmf{0}$, which can be handled numerically.
    If this problem is solved using the APG algorithm introduced in Algorithm~\ref{alg: APG}, the active set can be identified naturally as a byproduct of the optimization process.
    Once the active set is obtained, the KKT system can then be solved accordingly. 

    \vspace{-0.6ex}
    \subsection{Incorporating Direct Multiple-Shooting Scheme} \label{subsec: FDDP}
    In underactuated scenarios, obtaining a good initial rollout is generally challenging.
    FDDP \cite{FDDP} proposes a novel formulation that enables DDP computations even from dynamically infeasible state-control trajectories.
    A \textit{state gap} $\bar{\bmf{f}}$ is defined as:
    \begin{equation} \label{eq: gap}
        \bar{\bmf{f}}_{k+1} := \bmf{f}_k(\bmf{x}_k,\bmf{u}_k) \ominus \bmf{x}_{k+1} 
    \end{equation}
    where $\ominus$ is a \textit{difference} operator on the state manifold \cite{Box-FDDP}.
    This formulation treats the integrated state $\bmf{f}_k$ from $\bmf{x}_k$ and the state $\bmf{x}_{k+1}$ as independent quantities, and defines their discrepancy as the gap.
    Using this definition, the control law obtained from the backward pass can be modified, and the forward pass rule is given as:
    \begin{equation} \label{eq: FDDP-rollout}
    \begin{aligned}
        \mathbf{x}^*_0 &= \tilde{\mathbf{x}}_0 \oplus (\alpha - 1)\,\bar{\mathbf{f}}_0, \\
        \mathbf{u}^*_k &= \mathbf{u}_k + \alpha \mathbf{k}_k + \mathbf{K}_k(\mathbf{x}^*_k \ominus \mathbf{x}_k), \\
        \mathbf{x}^*_{k+1} &= \mathbf{f}(\mathbf{x}^*_k, \mathbf{u}^*_k) \oplus (\alpha - 1)\,\bar{\mathbf{f}}_{k+1}.
    \end{aligned}
    \end{equation}
    where $\alpha\!\in\!(0,1]$ and $\oplus$ is \textit{integrator} operator on the state manifold \cite{Box-FDDP}.
    Based on the definition of the state gap, \eqref{eq: FDDP-rollout} can be rewritten as $\bmf{x}_{k+1}^* = \bmf{x}_{k+1} \oplus \alpha\bar{\bmf{f}}_{k+1}$.
    The updated gap is then given by $\bar{\bmf{f}}_{k+1}^{\mathrm{updated}} = \bmf{f}_k(\bmf{x}_k,\bmf{u}_k)\ominus\bmf{x}_{k+1}^* = (1-\alpha)\,\bar{\bmf{f}}_{k+1}$.
    In other words, the gap decreases by a factor of $(1-\alpha)$ at each iteration, gradually steering the trajectory toward dynamic feasibility.
    Note that when $\alpha = 1$, \eqref{eq: FDDP-rollout} becomes identical to the classical DDP forward pass.
    Moreover, when $\bar{\bmf{f}}_{k+1} = \bmf{0}$, the update reduces to a backtracking step of the feedforward term scaled by $\alpha$.

    In practice, the initial iterations trade off the cost optimality to progressively improve dynamic feasibility.
    The step size $\alpha$ determines how conservatively the update proceeds in reducing the gap.
    Moreover, during backtracking, the effect of the gap is incorporated into the value function as a correction term \cite{Box-DDP}:
    \begin{equation} \label{eq: value-correction}
        V_{\bmf{x}_{k+1}} \leftarrow V_{\bmf{x}_{k+1}} + V_{\bmf{xx}_{k+1}} \bar{\bmf{f}}_{k+1}.
    \end{equation} 

    \subsection{Virtual Constraints and Boundary Shrinkage}
\begin{figure}[!t]
        \centering
        \includegraphics[width=0.95\columnwidth,
                trim=0 2ex 0 0,clip]
                {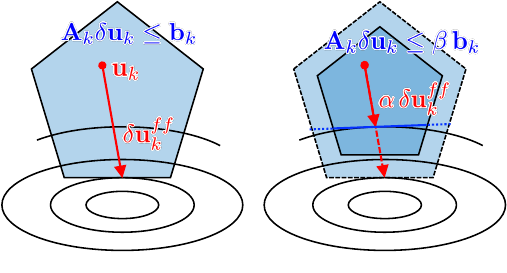}
        \caption{
            Backtracking in the presence of active constraints. 
            During backtracking, the feedforward step defines a virtual constraint (blue hyperplane), along which the feedback step evolves. 
            The original constraint set is contracted by a factor $\beta$, limiting the admissible region for the feedback component.
        }
        \label{fig: virtual-and-shrink}
    \end{figure}

    However, when FDDP backtracking is applied in the presence of active constraints, the hessian evaluated at the active boundary may lose its direct interpretability.
    At the same time, the resulting point does not correspond to an unconstrained local minimum, making it difficult to characterize within the standard framework.
    To address this, we introduce a \textit{virtual constraint}, interpreting the backtracked feedforward step as being constrained to lie on a hyperplane normal to the gradient:
    \begin{equation} \label{eq: virtual-constraint}
        \begin{aligned}
            \bmf{A}_k^* &\leftarrow \nabla F(\alpha\,\delta \mathbf{u}_k^{\mathrm{ff}})^\top \\
            \bmf{b}_k^* &\leftarrow \nabla F(\alpha\,\delta \mathbf{u}_k^{\mathrm{ff}})^\top (\alpha\,\delta \mathbf{u}_k^{\mathrm{ff}}),
        \end{aligned}
    \end{equation}
    when rewriting $\bmf{u}^*_k=\bmf{u}_k+\alpha\delta\mathbf{u}_k^{\mathrm{ff}}+\delta\mathbf{u}_k^{\mathrm{fb}}$.
    As a result, the feedback component $\delta\bmf{u}^{\mathrm{fb}}_k$ slides along this hyperplane, marked as a blue line in Fig.~\ref{fig: virtual-and-shrink}.
    
    \begin{figure}[!t]
        \centering
        \includegraphics[width=1.0\columnwidth]{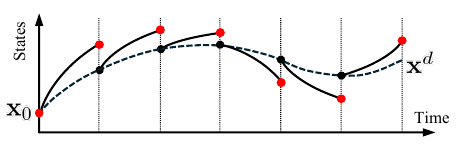}
        \caption{
            Illustration of a dynamically infeasible initial rollout. 
            The nominal trajectory (red) is generated by propagating the system dynamics from the desired trajectory (dashed line), resulting in a trajectory that is not dynamically feasible.
            Nevertheless, the trajectory remains bounded around the desired trajectory without diverging, even under prolonged underactuation.
        }
        \label{fig: initial-rollout}
    \end{figure}
    
    \begin{algorithm}[!t]
        \caption{APG-Based Control-Constrained DDP} \label{alg: ABCDDP}
        \textbf{Initialize} $c_\beta$$\in(0,1]$, $\bar{c}_{\alpha},\bar{c}_{\gamma}>1$, $\underbar{c}_{\alpha},\underbar{c}_{\gamma}<1$, $\gamma_0\ll 1$ \\
        Initial rollout by \eqref{eq: initialQP} \\
        \For{$i = 1,2,3, \cdots$}{
            gap computation by \eqref{eq: gap} \\
            \If{$\|\,\bar{\bmf{f}}\,\|_{\infty} \le \varepsilon_{gap}$}{
                $\bmf{f} \leftarrow \bmf{0}$
            }
            $\alpha \leftarrow 1$ \\
            \For{$i_b = 1,2,3, \cdots$}{
                    \For(\,\,\texttt{\small// Backward Pass}){$k = N-1,\cdots,0$}{
                    Update Taylor coefficients by \eqref{eq: value-correction} and \eqref{eq: Qdiffs} \\
                    Solve \eqref{eq: constrainedDDP} with $\delta\bmf{x}_k=\bmf{0}$ using Algorithm~\ref{alg: APG} \\ 
                    Identify $\bmf{k}_k$ and active-sets $\bmf{A}_k^*\delta\bmf{u}_k=\bmf{b}_k^*$\\
                    \If{$\textnormal{rows}(\bmf{A}_k^*)\neq0$ \textnormal{\textbf{and}} $\alpha<1$}{
                        Update $\bmf{A}_k^*$ and $\bmf{b}_k^*$ by \eqref{eq: virtual-constraint} 
                    }
                    Solve KKT equation \eqref{eq: KKT-DDP} to obtain $\bmf{K}_k$  \\
                    Update value function \eqref{eq: value-function}
                } 
                $\beta \leftarrow 1$ \\
                \For{$i_c = 1,2,3, \cdots$}{
                    Forward pass by \eqref{eq: FDDP-rollout} and \eqref{eq: shrink} \\
                    Calculate expected improvement \eqref{eq: expected-improvement} \\
                    \If{\eqref{eq: stop-criteria} fails}{
                        $\beta \leftarrow c_{\beta}\cdot\beta$
                    }
                }
                \If{\eqref{eq: stop-criteria} holds}{
                    $\alpha \leftarrow \textnormal{max}(\bar{c}_{\alpha}\cdot\alpha,1)$ \\
                    \If{$\alpha = 1$}{
                        $\gamma \leftarrow \underbar{c}_{\gamma}\cdot\gamma$
                    }
                    \textnormal{\textbf{break}}
                }
                $\alpha \leftarrow \underbar{c}_{\alpha}\cdot\alpha$ \\
                \If{$\alpha<1/2$}{
                    $\gamma \leftarrow \bar{c}_{\gamma}\cdot\gamma$
                }
                Update $\bmf{x}_k\leftarrow\bmf{x}_k^*$, $\bmf{u}_k\leftarrow\bmf{u}_k^*$ for $\forall k$
            }

            \If{$|\ell_{i}-\ell_{i+1}|^2 < \varepsilon_{cost}$ \,\textnormal{\textbf{and} $\|\,\bar{\bmf{f}}\,\|_{\infty} = 0$ }  }{
                \textbf{break}
            }
        }
    \end{algorithm}

    Meanwhile, the feedforward component in \eqref{eq: backpass-solution} is guaranteed to satisfy the constraints, as it is obtained numerically using APG.
    However, during the forward pass, the feedback component may violate the constraints, which cannot be anticipated during the backward pass.
    To address this, a projection step $P_{C_k}(\delta\bmf{u}_k^*)$ is applied prior to each integration.
    However, the virtual constraint, which is obtained with the backtracking step $\alpha\,\delta\bmf{u}_k^{\mathrm{ff}}$, does not carry a direct physical interpretation, 
    and it is therefore desirable to further limit the search region. To this end, the constraint boundary is contracted linearly around $\bmf{u}_k^{\mathrm{ff}} = \bmf{u}_k+\alpha\,\delta\bmf{u}_k^{\mathrm{ff}}$.
    The contracted constraint is then given as follows:
    \begin{equation} \label{eq: shrink}
        C'_k := \{ \delta\bmf{u}_k^{\mathrm{fb}} \mid \bmf{A}_k\delta\bmf{u}_k^{\mathrm{fb}} - \beta\mathbf{b}_k^{\mathrm{ff}} \leq \bmf{0} \}
    \end{equation}
    where $\beta \in (0,1]$. 
    When $\beta = 1$, the constraint reduces to the original friction pyramid $C_k$.
    The parameter $\beta$ controls how close the constraint boundary is to the center.
    It can be inter-preted as another form of backtracking, complementary to $\alpha$.

    \subsection{Dynamically Infeasible Initial Rollout}
    For the initial rollout of DDP, the nominal trajectory is generated using the following PD-like control law:
    \begin{equation} \label{eq: initialQP}
        \begin{aligned}
            \bmf{x}^{\mathrm{nom}}_{k+1} &:= \bmf{f}_k(\bmf{x}^{d}_{k},\bmf{u}^{\mathrm{nom}}_k),\quad\bmf{x}^{d}_0:=\tilde{\bmf{x}}_0 \\
            \mathbf{u}^{\mathrm{nom}}_k &:= P_{C_k}\left( \operatorname*{arg\,min}_{\mathbf{u}}\;\frac{1}{2}\|\mathbf{G}\mathbf{u}-\mathbf{h}\|^2+\frac{\rho}{2}\|\mathbf{u}\|^2 \right) \\
        \end{aligned}    
    \end{equation}
    where
    $\mathbf{G}\mathbf{u}=\mathbf{h}$ constitutes a Newton-Euler equation in matrix form as in \cite{2017HyQ}.
    For the $6\times1$ generalized acceleration $\bmf{a}$, position $\bmf{p}$ and velocity $\bmf{v}$ of a rigid body, the desired acceleration is formulated as  
    $\mathbf{a}^d = \mathbf{K}_P(\mathbf{p}^{d}_{k+1}\ominus\mathbf{p}^{d}_{k}) + \mathbf{K}_D(\mathbf{v}^{d}_{k+1}\ominus\mathbf{v}^{d}_{k})$.
    The error is constructed using the difference between consecutive desired states.
    If the trajectory were propagated from $\bmf{x}_0$ using a single-shooting scheme, this kind of formulation would effectively provide only a one-step update, which would inevitably lead to divergence under underactuated conditions.
    By incorporating FDDP, however, the state and control trajectories are allowed to be dynamically infeasible during intermediate iterations.
    This enables the construction of an initial trajectory that, although seemingly inconsistent, remains bounded around the desired trajectory, as illustrated in Fig.~\ref{fig: initial-rollout}.

    \subsection{Expected Improvement}
    Even though the backtracking step is intended to improve dynamic feasibility, 
    excessive compromise in cost reduction should be avoided, and thus an appropriate criterion is needed.
    We adopt the stopping criterion for backtracking proposed in \cite{Box-FDDP}.
    The expected improvement along the feasibility-driven direction is approximated up to second order as follows:
    \begin{equation} \label{eq: expected-improvement}
        \Delta J(\alpha) = \Delta_1 \alpha + \frac{1}{2}\Delta_2 \alpha^2
    \end{equation}
    where the coefficients are defined using the derivatives obtained during backward pass as:
    \begin{equation}
    \begin{aligned}
        \Delta_1 &= \textstyle\sum_{k=0}^{N-1} 
        \mathbf{k}_k^{\top} \mathbf{Q}_{\bmf{u}_k}
        + \bar{\mathbf{f}}_k^{\top}
        \left( 
        V_{\bmf{x}_{k+1}} - V_{\bmf{xx}_{k+1}}\delta\mathbf{x}_k 
        \right),
        \\[0.5ex]
        \Delta_2 &= \textstyle\sum_{k=0}^{N-1}
        \mathbf{k}_k^{\top}\mathbf{Q}_{\bmf{uu}_k}\mathbf{k}_k
        + \bar{\mathbf{f}}_k^{\top}
        \left( 
        2V_{\bmf{xx}_{k+1}}\delta\mathbf{x}_k
        - V_{\bmf{xx}_{k+1}}\bar{\mathbf{f}}_k 
        \right),
    \end{aligned}
    \end{equation}

    \noindent 
    \begin{equation} \label{eq: stop-criteria}
        \ell'-\ell\leq
        \begin{cases}
        b_1 \Delta J(\alpha) & \text{if } \Delta J(\alpha) \le 0, \\
        b_2 \Delta J(\alpha) & \text{otherwise}
        \end{cases}
    \end{equation}
    where $\ell'$ means the cost value updated during the forward pass and $b$'s are adjustable parameters.
    This condition allows for moderate increases in the objective value, preventing overly conservative updates that would otherwise limit the step size.
    At the same time, it enables the iterations to focus on improving dynamic feasibility rather than reducing the cost during the initial feasibility-driven phase.

    \begin{table}[!t]
        \centering
        \caption{System Parameters}
        \label{table: system-parameters}
        \begin{tabular}{ccc|ccc}
        \hline \noalign{\vspace{0.25ex}}
        Parameter & Value & Unit & Parameter & Value & Unit \\
        \hline\hline \noalign{\vspace{0.25ex}}
        
        $m_{body}$ & 37.5 & $\rm{kg}$ & Body length & 0.6448 & $\rm{m}$ \\
        $I_{xx}$ & 0.7 & $\rm{kg \cdot m^2}$ & Body width & 0.119 & $\rm{m}$ \\
        $I_{yy}$ & 2.8 & $\rm{kg \cdot m^2}$ & Scap length & 0.1377 & $\rm{m}$ \\
        $I_{zz}$ & 3.3 & $\rm{kg \cdot m^2}$ & Hip length & 0.35 & $\rm{m}$ \\
        $\mu$ & 0.5 & & Knee length & 0.4136 & $\rm{m}$ \\
        
        \hline
        \end{tabular}
        
    \end{table}

    \begin{figure}[!t]
        \centering
        \includegraphics[width=0.7\columnwidth]{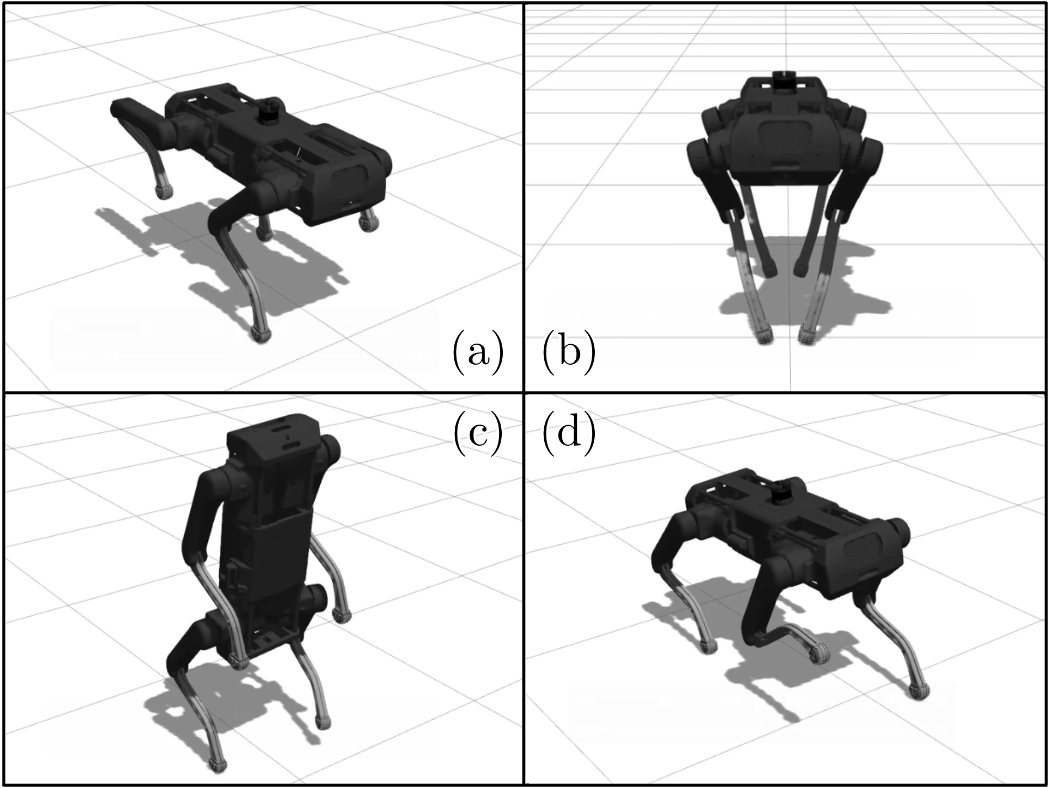}
        \caption{
            Four representative motions: (a) static two-leg standing, (b) slow catwalk, (c) upright walking, and (d) high-speed running.
        }
        \label{fig: test-motions}
    \end{figure}

    \begin{table}[!t]
        \centering
        \caption{MPC Parameters} \label{table: MPCparameters}
        \begin{tabular}{ccccc}
        \hline \noalign{\vspace{0.5ex}}
         & Two-Leg & Catwalk & Upright Walk & Running \\
        \hline\hline \noalign{\vspace{0.25ex}}
        $W_{p_x}$ & 2e5 & 1e3 & 2.5e3 & 1e4 \\
        $W_{p_y}$ & 2e5 & 1e3 & 2.5e3 & 1e4 \\
        $W_{p_z}$ & 2e5 & 1e3 & 2.5e3 & 2e5 \\
        \hline \noalign{\vspace{0.25ex}}
        $W_{v_x}$ & 1 & 5 & 1 & 1e3 \\
        $W_{v_y}$ & 1 & 5 & 1 & 1e3 \\
        $W_{v_z}$ & 1 & 1 & 1 & 50 \\
        \hline \noalign{\vspace{0.25ex}}
        $W_{R_x}$ & 12 & 10 & 150 & 50 \\
        $W_{R_y}$ & 36 & 10 & 150 & 50 \\
        $W_{R_z}$ & 10 & 10 & 150 & 50 \\
        \hline \noalign{\vspace{0.25ex}}
        $W_{\omega_x}$ & 0.1 & 1 & 0.1 & 1.5e-2 \\
        $W_{\omega_y}$ & 0.15 & 1 & 0.1 & 1.5e-2 \\
        $W_{\omega_z}$ & 0.05 & 0.1 & 0.1 & 1.5e-2 \\
        \hline \noalign{\vspace{0.25ex}}
        $W_{u_x}$ & 1e-4 & 1e-5 & 1e-2 & 1e-3 \\
        $W_{u_y}$ & 1e-4 & 1e-5 & 1e-2 & 1e-3 \\
        $W_{u_z}$ & 1e-4 & 1e-5 & 1e-4 & 2e-4 \\
        \hline \noalign{\vspace{0.25ex}}
        $W_{\Delta u_x}$ & 5e-4 & 0 & 1e-2 & 0 \\
        $W_{\Delta u_y}$ & 5e-4 & 0 & 1e-2 & 0 \\
        $W_{\Delta u_z}$ & 1e-3 & 0 & 1e-4 & 0 \\
        \hline \noalign{\vspace{0.25ex}}
        $T_{sw}$ & N/A & 0.6 & 0.2 & 0.08 \\
        $T_{st}$ & N/A & 1.0 & 0.24 & 0.10 \\
        $N_{hor}$ & 15 & 15 & 15 & 15 \\
        $\Delta t_{pred}$ & 0.04 & 0.02 & 0.04 & 0.02 \\
        $f_{MPC}$ & 25 & 50 & 25 & 50 \\
        Avg. solve time & 0.0052 & 0.0012 & 0.0055 & 0.0013 \\
        Max. solve time & 0.024 & 0.0032 & 0.0187 & 0.0084 \\
        $u_{z,max}$ & 666 & 666 & N/A & N/A \\
        $u_{z,min}$ & 50 & 10 & 50 & 10 \\
        \hline
        \end{tabular}
        \begin{minipage}{0.92\columnwidth}
            \vspace{1ex}\footnotesize
            \textit{Note:} All time units are in [sec] and force units are in [N]. 
            $T_{sw}$ and $T_{st}$ denote swing and stance durations, respectively. 
            $N_{hor}$ is the MPC prediction horizon, $\Delta t_{pred}$ is the prediction time step, and $f_{MPC}$ is the MPC loop frequency in [Hz].
        \end{minipage}        
    \end{table}

    \vspace{-1.2ex}
    \subsection{Regularization}
    If \eqref{eq: stop-criteria} is not satisfied, regularization is applied to the hessian matrices:
    \begin{equation}
    \begin{aligned}
        Q_{\bmf{uu}_k} &\leftarrow Q_{\bmf{uu}_k} + \gamma \cdot \mathbb{I}  \\
        V_{\bmf{xx}_{k+1}} &\leftarrow V_{\bmf{xx}_{k+1}} + \gamma \cdot \mathbb{I} 
    \end{aligned}
    \end{equation}
    where $\gamma>0$ and $\mathbb{I}$ is the appropriate identity matrix. 
    The complete algorithm of the ABC-DDP is summarized in Algorithm~\ref{alg: ABCDDP}.

\vspace{-1.2ex}
\section{Results} \label{sec: Results}
    This section presents experimental results of the proposed ABC-DDP framework. 
    The robot specifications, including size and mass properties, are summarized in Table~\ref{table: system-parameters}.
    Four representative motion tasks are evaluated, as shown in Fig.~\ref{fig: test-motions}: static two-leg standing, slow catwalk, upright walking, and high-speed running. 
    In the catwalk motion, each foot is placed approximately 0.18~m closer to the centerline to reduce the support polygon. 
    The MPC parameters used for each task are listed in Table~\ref{table: MPCparameters}.
    The manuscript focuses on detailed results for the static two-leg standing task, while additional results including other motions are provided in the supplementary video.

    \vspace{-1.2ex}
    \subsection{Offline Computation Results}
    \subsubsection{Effect of Incorporating FDDP}
    Fig.~\ref{fig: offline-compare} (a),(b) shows the effect of incorporating FDDP, as discussed in Section~\ref{subsec: FDDP}. 
    The problem is solved with a slightly perturbed initial state in a two-leg standing configuration, using $\Delta t=0.04$s and $N=200$. 
    Since generating a feasible initial rollout without FDDP is difficult, we compare the full FDDP implementation with a baseline where FDDP is applied only once at the first iteration with $\alpha=1$. 
    The full implementation achieves improved stability and converges to a lower objective value. 
    
    \subsubsection{Constraint Handling with APG vs.\ QP}
\begin{figure}[!t]
        \centering
        \includegraphics[width=1.0\columnwidth]{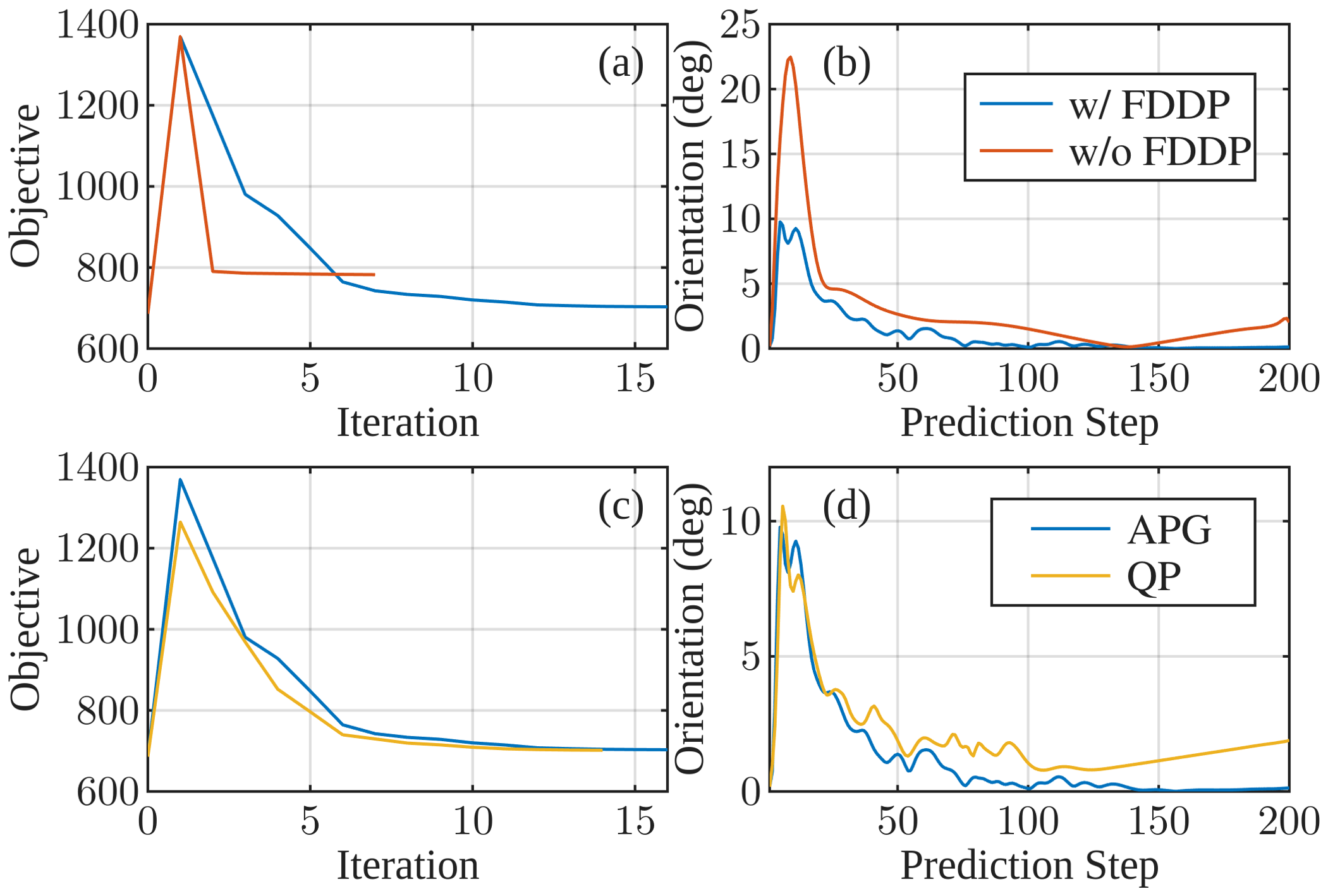}
        \caption{
            Comparison of convergence behavior and constraint handling performance. 
            (a),(b) Effect of incorporating FDDP, showing improved convergence and stability. 
            (c),(d) Comparison between APG and QP for solving the constrained subproblem, where APG achieves comparable convergence with improved robustness.  
            Note that APG achieved approximately 2.5$\times$ faster computation than QP.
        }
        \label{fig: offline-compare}
    \end{figure}

    Fig.~\ref{fig: offline-compare} (c),(d) compares APG and QP for solving the constrained subproblem in an offline setting. 
    Experiments were performed in MATLAB 2025a on an Intel Core i5-6600 @3.30GHz desktop with 16GB RAM, using \texttt{quadprog} for QP.  
    Both methods showed a similar number of iterations, but APG is significantly faster since QP requires KKT matrix inversion at each iteration.
    For five runs of a 200-step problem, APG required on average 16 iterations and 10.54~s, whereas QP required 14 iterations and 26.10~s. 
    In addition, QP occasionally misidentifies active constraints, resulting in inferior solution trajectories. 
    These results indicate that APG is well suited for solving simply constrained problems with reliable active-set identification.

    \begin{figure}[!t]
        \centering
        \includegraphics[width=1.0\columnwidth]{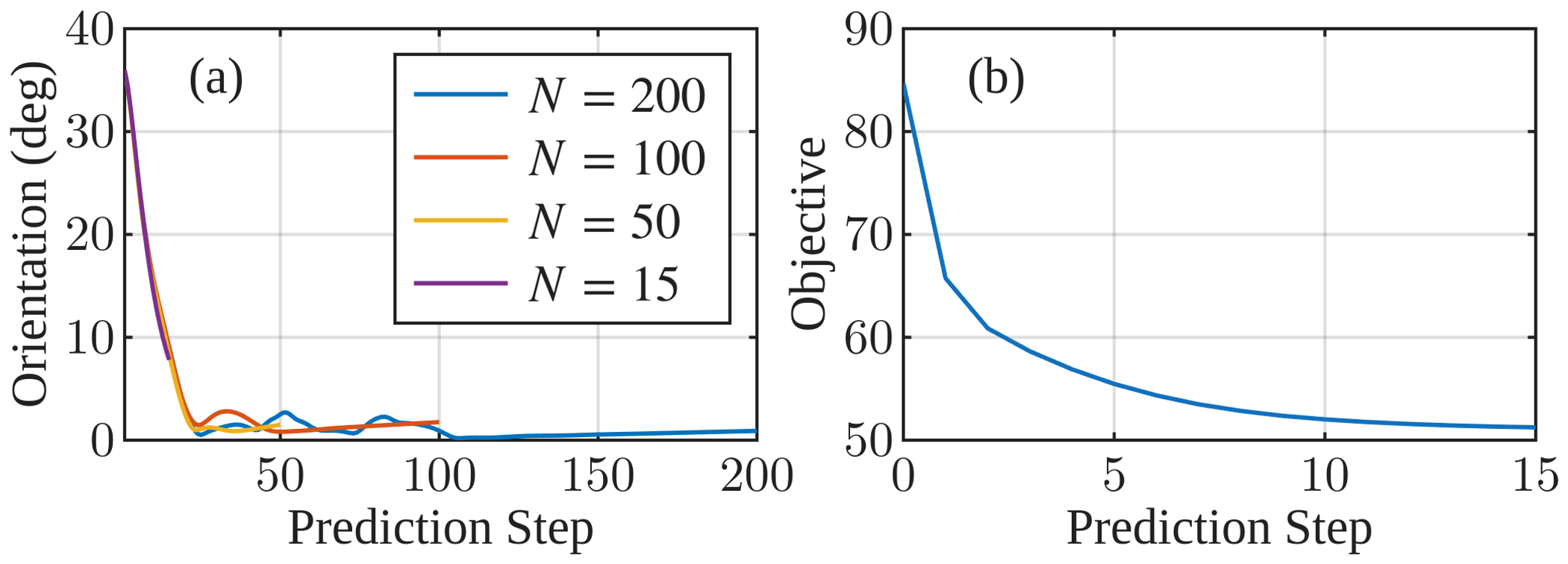}
        \caption{
            Effect of the prediction horizon. 
            (a) Similar orientation trajectories are obtained for different horizon lengths, indicating robustness to horizon reduction. 
            (b) Objective value for $N = 15$ under a receding-horizon scheme shows consistent decrease over the prediction step, indicating stable MPC behavior even with a short horizon.
        }
        \label{fig: offline-horizon}
    \end{figure}

    \begin{figure}[!t]
        \centering
        \includegraphics[width=1.0\columnwidth]{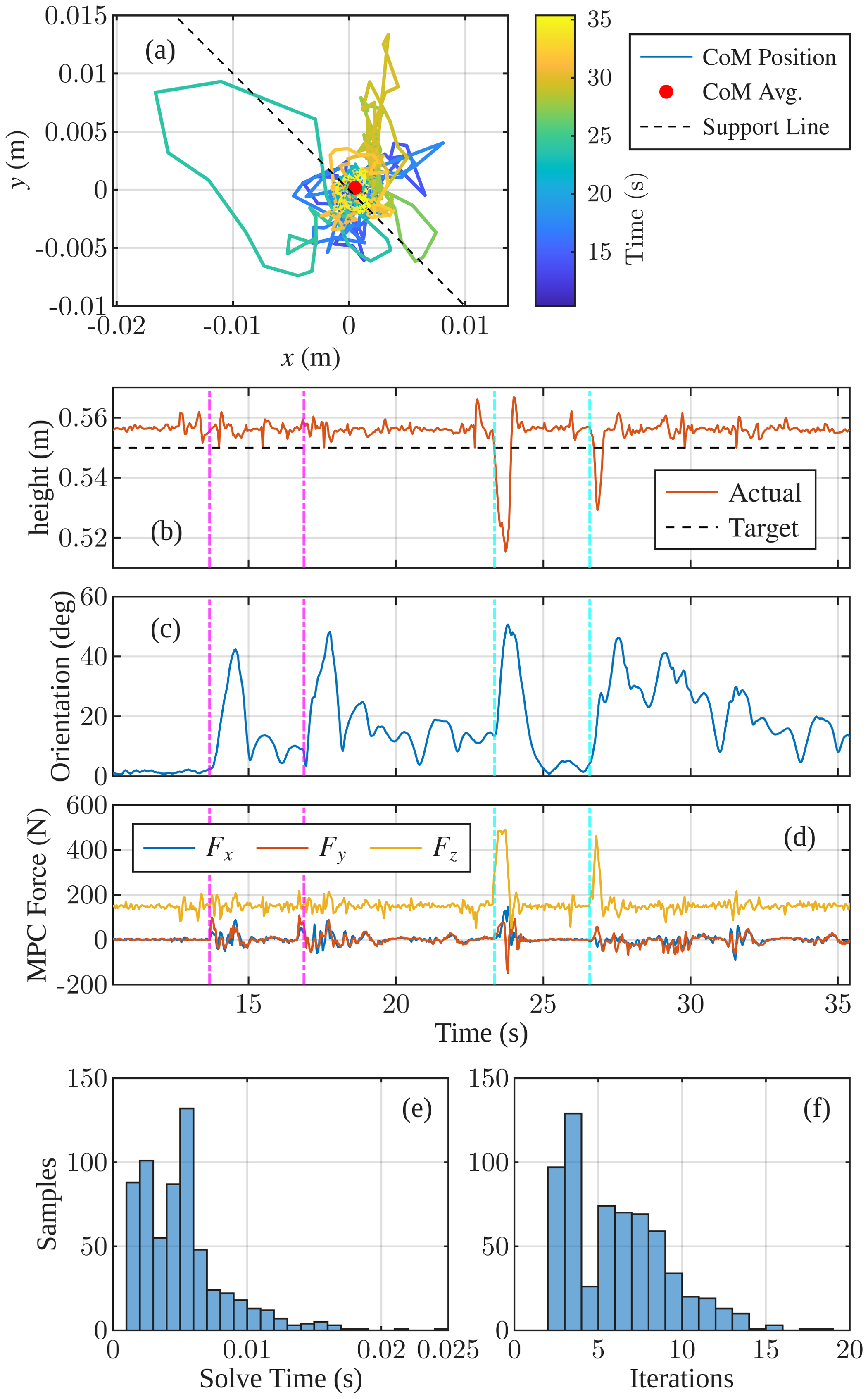}
        \caption{
            Experimental results of two-leg standing under external disturbances. 
            (a) Center of mass (CoM) trajectory in the $xy$ plane, with color indicating time and the red marker denoting the average CoM position. 
            (b) Height, (c) orientation, and (d) MPC force responses (front-right leg). External impulses are applied at the indicated time instants (magenta: lateral, cyan: vertical). 
            (e),(f) Histograms of solve time and iteration count, showing real-time performance within the sampling period ($40 \mathrm{ms}$).
        }
        \label{fig: two-leg-disturbance}
    \end{figure}
    \setlength{\textfloatsep}{8pt}

    \subsubsection{Effect of the Prediction Horizon}
    Fig.~\ref{fig: offline-horizon} (a) shows that similar trajectories are obtained even as the prediction horizon is reduced from several hundred steps to $N=15$, indicating robustness to horizon length. 
    For $N=15$, a receding-horizon scheme is applied, and the objective value over time is shown in (b). 
    Despite an initial tilt of approximately $35^\circ$, the objective decreases consistently, indicating stable convergence. 
    This suggests that accurate short-horizon MPC can still generate complex motions.

    \subsection{Real-time Simulation}
    The simulation runs on a desktop, while the controller is executed on a separate SBC (GENE-KBU6, Intel Core i7-6600U CPU @2.60GHz, 8GB memory) via ROS communication.

    \subsubsection{Static Two-leg Standing under External Disturbances} \label{subsubsec: two-leg}
    External forces corresponding to approximately $25\%$ of the robot weight are applied near the shoulder for $0.2$s ($-100$~N in $y$, $-150$~N in $z$). 
    As illustrated in Fig.~\ref{fig: SRBD}, the robot maintains balance by generating rotational motion similar to a spinning top. 
    The resulting precession and nutation effects, arising from the nonlinear term in rotational dynamics, a.k.a. $\bmf{\omega}\times\bmf{I}\bmf{\omega}$, keep the body near the support line, as shown in Fig.~\ref{fig: two-leg-disturbance} (a), with an average CoM position of $[0.5268,\,0.1639]$~mm. 
    Fig.~\ref{fig: two-leg-disturbance} (b)--(d) show height, orientation, and MPC force responses, where disturbances are applied at the indicated time instants. 
    The robot rapidly recovers its posture after each disturbance. 
    The histograms in (e),(f) show solve time and iteration count. 
    With a control period of $40$~ms ($25$~Hz), the maximum solve time is $24$~ms, leaving sufficient computational margin, and the iteration count remains below the maximum limit set as 50.

    Box-FDDP is one of the most successful DDP approaches for legged robots, 
    where box constraints handle joint-torque limits while friction constraints generally rely on penalty-based methods. 
    Thus, penalty-based FDDP provides the closest existing counterpart to our formulation, and a qualitative comparison is provided in the supplementary video 
    using $F_{\textnormal{penalty}}=\frac{1}{2}W_{\mathrm{fric}}\|\max(\mathbf{A}_k\delta\mathbf{u}_k-\mathbf{b}_k,\mathbf{0})\|^2$. 
    The comparison favors our method, consistent with \cite{XieConstrainedDDP}, where explicit constraint handling also outperformed penalty-based handling.

    \subsubsection{Effect of Friction Coefficient}
\begin{figure}[!t]
        \centering
        \includegraphics[width=1.0\columnwidth]{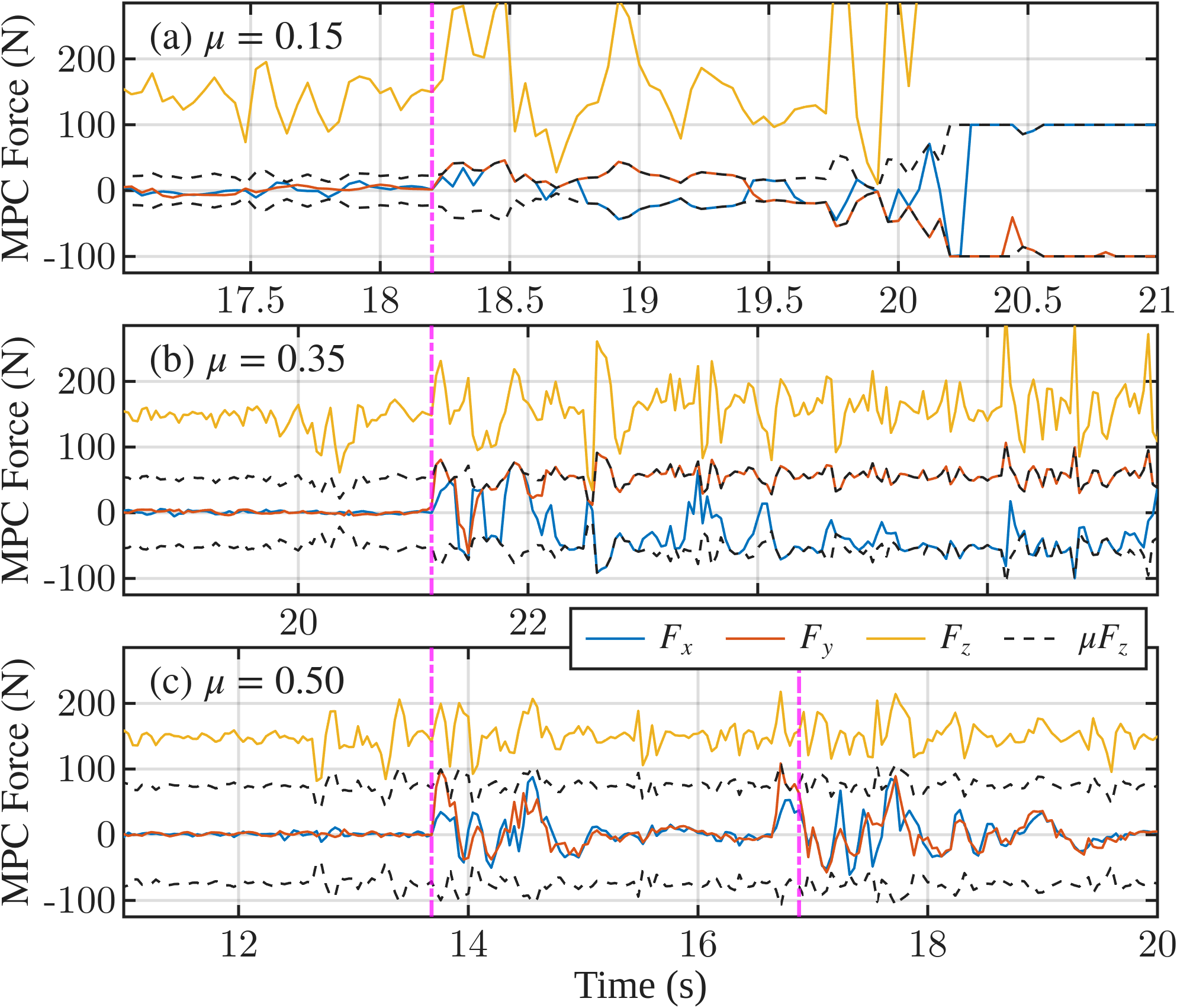}
        \caption{
            Comparison of the front-right leg contact forces for different friction coefficients ($\mu=0.15,0.35,0.50$).
            The magenta dashed lines indicate the instants at which the lateral disturbances are applied, while the dashed black lines represent the friction-pyramid limits.
        }
        \label{fig: two-leg-mu}
    \end{figure}

    The effect of the friction coefficient used in the MPC was investigated in the two-leg standing task. While the experiment in Section~\ref{subsubsec: two-leg} was conducted with $\mu=0.5$, the same experiment was repeated under $\mu=0.15, 0.35, 0.5$ using an identical lateral impulse.
    Fig.~\ref{fig: two-leg-mu} shows the resulting MPC forces. For $\mu=0.15$, the available lateral force is highly restricted, causing the controller to rely primarily on increasing $F_z$ and ultimately fail to recover the balance. 
    With $\mu=0.35$, the contact forces consistently reach the friction-pyramid limits, and the robot is still unable to regain the nominal posture.
    In contrast, $\mu=0.5$ enables successful recovery from successive disturbance events without violating the contact constraints.
    The remaining experiments therefore employ $\mu=0.5$. These results further confirm that the proposed MPC properly handles the contact constraints.

    \subsubsection{Command Tracking in Two-leg Standing}
\begin{figure}[!t]
        \centering
        \includegraphics[width=1.0\columnwidth]{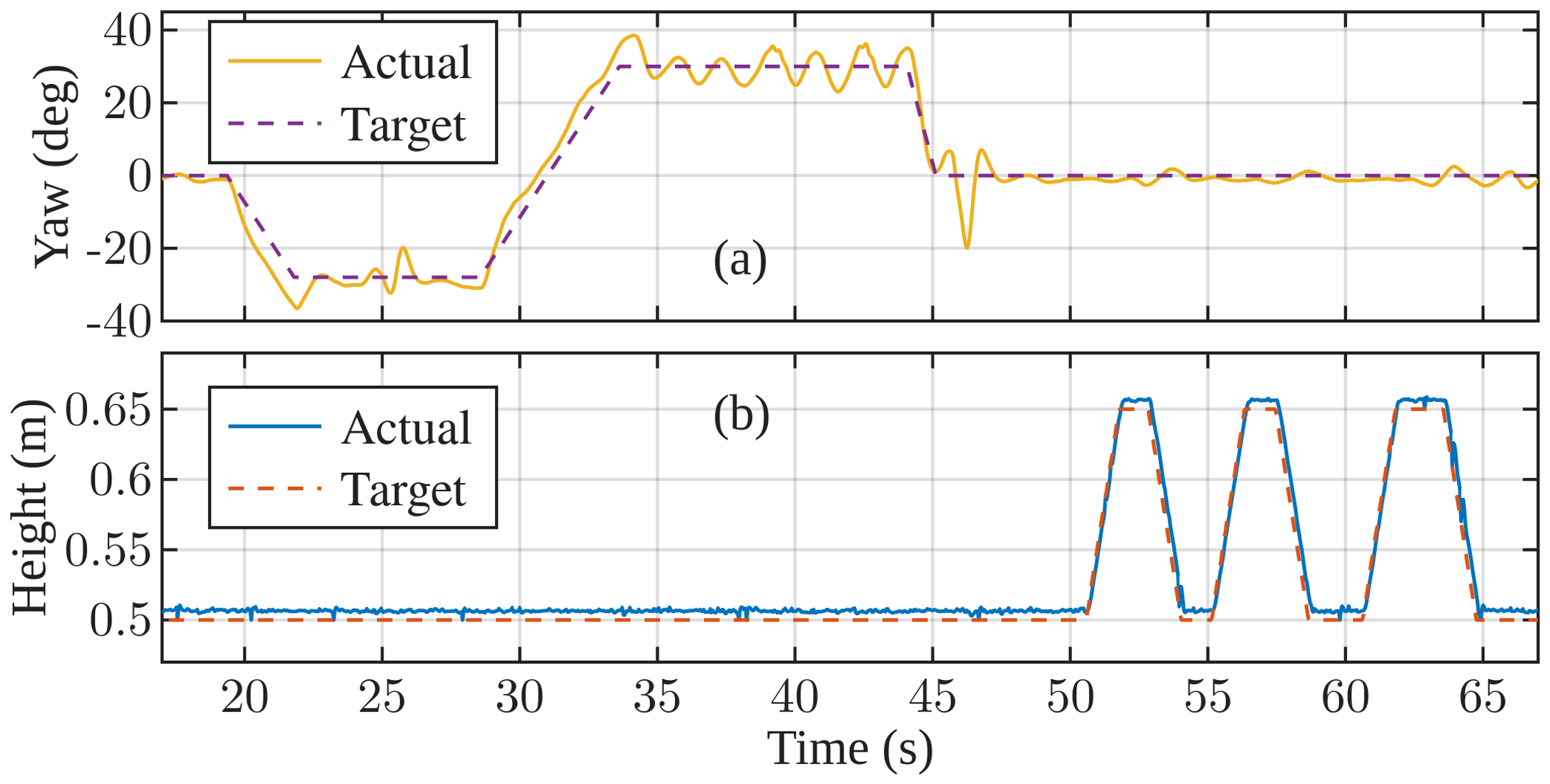}
        \caption{
            Command tracking during two-leg standing. 
            (a) The yaw response follows the reference even under underactuation, with oscillatory behavior. 
            (b) Height follows the reference in a consistent and stable manner.
        }
        \label{fig: two-leg-cmd}
    \end{figure}

    Fig.~\ref{fig: two-leg-cmd} shows command tracking under underactuated conditions. 
    Despite limited actuation, the controller maintains responsiveness to the commanded posture. 
    Additional visual results, including the other tasks, are provided in the supplementary video.

    \setlength{\textfloatsep}{18pt}

\section{Conclusion} \label{sec: Conclusion}
    This paper presented an APG-based control-constrained DDP framework for real-time model predictive control of underactuated legged robots. 
    By integrating projection-based constraint handling, KKT-based formulations, and an accelerated first-order method, the proposed approach efficiently solves control-constrained DDP subproblems while avoiding repeated matrix inversions. 
    The introduction of a virtual constraint further enables consistent integration with a feasibility-driven multiple-shooting scheme, allowing stable trajectory optimization even from dynamically infeasible initializations.
    The effectiveness of the proposed method was demonstrated through various tasks, including static two-leg standing under external disturbances and dynamic locomotion behaviors, all within a unified real-time MPC framework.

    Although dynamics hessian and non Gauss--Newton components in \eqref{eq: Qdiffs} were also explored, their performance was found to degrade as the step size increases, 
    likely due to distortion from higher-order terms in Taylor expansion. 
    This suggests that selectively incorporating informative second-order components, beyond conventional Gauss--Newton approximations, could further improve performance. 
    In addition, adaptive weight scheduling across different tasks remains an important topic for enhancing robustness and generality.

\bibliographystyle{bibfiles/IEEEtran.bst} 
\bibliography{bibfiles/bibfile.bib}

\end{document}